# The impact of external innovation on new drug approvals: A retrospective analysis

Xiong Liu[1], Craig E. Thomas[2], Christian C. Felder[3]

1. Advanced Analytics and Data Sciences, Eli Lilly and Company, Indianapolis, IN 46285
2. Lilly Research Laboratories, Eli Lilly and Company, Indianapolis, IN 46285
3. Discovery Research, Karuna Pharmaceuticals Inc, Boston, MA 02110

*Corresponding author*:
Xiong Liu (xliu09@gmail.com)
Currently at Novartis

## Cite:

# Abstract

Pharmaceutical companies are relying more often on external sources of innovation to boost their discovery research productivity.  However, more in-depth knowledge about how external innovation may translate to successful product launches is still required in order to better understand how to best leverage the innovation ecosystem. We analyzed the pre-approval publication histories for FDA-approved new molecular entities (NMEs) and new biologic entities (NBEs) launched by 13 top research pharma companies during the last decade (2006-2016). We found that academic institutions contributed the majority of pre-approval publications and that publication subject matter is closely aligned with the strengths of the respective innovator. We found this to also be true for candidate drugs terminated in Phase 3, but the volume of literature on these molecules is substantially less than for approved drugs. This may suggest that approved drugs are often associated with a more robust dataset provided by a large number of institutes. Collectively, the results of our analysis support the hypothesis that a collaborative research innovation environment spanning across academia, industry and government is highly conducive to successful drug approvals.

# Keywords

Drug discovery; New molecular entities; New biologic entities; External innovation; Innovation ecosystem; Text mining; Natural language processing

# 1. Introduction

R&D productivity is a continuing challenge for the pharmaceutical industry [1]. Previous research examined how investment in R&D may affect product launches [2,3]. These studies showed that total R&D expenditure was proportional to the number of new molecular entities (NMEs) and new biologic entities (NBEs) launched from research-based pharmaceutical companies [2]. However, the efficiency of new launches as measured by the number of new FDA approved drugs per unit R&D expenditure has been declining for decades for the entire pharmaceutical industry [3]. The primary causes of decline in efficiency include:  blockbusters that become generic, more rigorous regulatory requirements, greater demand on improved safety and health outcomes through patient tailoring, poor lead discovery relying on brute force screening, and a poor understanding of less obvious factors affecting return on R&D investment [3]. To improve R&D effectiveness, the pharmaceutical industry has exploited many R&D models such as increased mergers and acquisition, establishing research hubs across the globe, and shifting from primary care small molecule medicines to specialty medicines and biologics targeting niche diseases with high unmet medical need [4]. More recently, pharmaceutical companies have been exploring less tested external innovation models to supplement internal discovery to generate value [5]. Representative models include risk-sharing partnerships between pharma-academic institutions [6,7], pharma-government partnerships [8], pharma peer agreements [5], and open innovation including crowd sourcing [9,10,11]. To date, no detailed analysis has been published to support which of these models may provide improvements in R&D efficiency which is likely due to the paucity of data available for these relatively new methods of improving R&D efficiency.

Several previous studies have looked at the association between publication characteristics and drug approvals. One group concluded that, after the year 2000, academics contributed around half of the first publications on approved NMEs and that on average more than five years elapsed between the first publication and the launch [12]. However, this study focused only on the first paper published related to the NME and did not analyze the full research volume that existed between first publication and NME approval, nor the source of those contributions as judged by primary author affiliation. We posited that a more robust analysis using large data sets could provide deeper insight as to key success factors influencing drug approval and launches. We therefore collected and analyzed FDA approval data during the last decade (2006-2016) and focused on pharmaceutical companies that invest heavily in R&D [2] as their approvals account for a reasonable percentage of total drug approvals [13]. We analyzed the approved NMEs and NBEs as a collective, as well as at the individual company level, using a literature text mining method that allowed examination of very large publication datasets [14]. To identify areas of focus of early research activity, we extracted from PubMed only those research manuscripts on approved drugs published prior to their approval. The publications had to include the drug name/synonym, which differs from other studies that have evaluated publications linked to drug targets and not necessarily the drug itself. We examined characteristics such as total research volume prior to the FDA approval, the time elapsed between first publication and approval, research contributor type based on analysis of primary institutions, and research content or emphasis by using major Medical Subject Headings (MeSH) term analysis. This same strategy was applied to failed drugs which we defined as those terminated in/after Phase 3 by the same 13 pharma during the 2006-2016 timeframe. When stratified by company or therapeutic area, our analysis of approved drugs revealed relatively consistent patterns of publication characteristics associated with successful FDA approvals. The publication patterns for failed drugs differed modestly on a few dimensions but the body of literature associated with them was significantly less than for approved drugs. Together, the analysis supports the hypothesis that enhanced investment and collaboration in the external academic innovation environment is likely to contribute to more efficient discovery and approval of innovative medicines.

## 2. Methods

In this paper, we focused on the NME and NBE launches of 13 large, research-based pharmaceutical companies [2] during 2006-2016 and studied the research contributions that were associated with these drug launches prior to their approval. The NME and biologic launch information was obtained from the FDA website [13]. Given that mergers and acquisitions occurred during the decade, subsidiaries are incorporated into the parent companies for analysis; for example, Genentech and Roche are considered as one institution.

In total, there were 115 total drugs approved (see Supplementary table 1). Two drugs were approved twice - alglucosidase alfa (biologic) which was approved in 2006 under 'Myozyme' and 2010 under 'Lumizyme' [15], and Tocilizumab (biologic) which was approved in 2010 for intravenous injection and 2013 for subcutaneous injection [16]. Thus, the analysis was conducted for 113 unique drugs.

Based on the approved indications for those 113 NMEs and biologics, we classified the drugs into 9 general therapeutic areas, including oncology, cardiovascular disease, endocrinology, immunology, infectious disease, neuroscience, respiratory disease, ophthalmology and rare diseases (*e.g.,* Pompe disease and Gaucher disease).

PubMed was used as the data source for identifying research papers that were published before the launch dates of the selected NMEs or biologics. The generic name, trade name and drug synonyms for each NME or biologic were searched in the PubMed database using the I2E text mining tool [17]. Specifically, we used the MeSH and NCI Thesaurus ontologies in I2E to find relevant concepts and synonyms given a generic or trade name. For combination drugs, we searched brand names as well as individual drug names in a phrase with 3 or less words between them. This I2E search strategy provided a rigorous and extensive analysis of the publicly available literature. We also took steps to remove publications we considered irrelevant to drug approvals. First, we eliminated papers published after the drug approval date. Second, papers labelled by PubMed as review or non-research articles were omitted as they do not contribute directly to the scientific validation of the target and thus in our judgement are less likely to be influential in a drug's approval. Papers that do not include any affiliation information were also not included as author affiliation was an essential component of our analysis.

For research contributor analysis, we extracted the primary authoring institutions for each publication and classified the organization into different categories using the organization ontology provided by I2E [17]. The major categories included four groups: the top 13 pharma, other pharma and biotech companies, academic organization (including universities, hospitals and medical institutes) [18], and government agencies such as the NIH. Those organizations that are not included in the previous list (e.g., private practices) are grouped into an "other" category. To study the content of the papers, we searched for MeSH major headings of each papers [19], analyzed the frequency of each MeSH term, and thereby identified the top 15 most frequent terms. We used these top 15 terms to represent the research content of our validated papers in order to best define the research pattern/type exemplified in each publication.

For comparison with approved NMEs and biologics, we also analyzed candidate drugs from the same top 13 pharma that failed at Phase 3 during 2006 to 2016. We queried the PharmaProjects database [20] and identified a list of 39 original drugs that were discontinued or suspended (see Supplementary table 2) and then performed the same analyses of those failed drugs as was done for the approved drugs.

## 3. Results
### 3.1 Approval characteristics
Between 2006 and 2016 the FDA approved a total of 251 NMEs and 67 biologics. In this paper, we focused on drug approvals from 13 top pharma and studied the publication history that preceded the drug launches. This group of pharmaceutical companies contributed 82 NME launches and 33 NBE launches (31 unique NBEs) providing approximately 33% and 50% of the total launches, respectively.

Figure 1 shows the number and relative percentage of NME and biologic approvals for these 13 companies. Novartis, Pfizer, and GSK generated the most launches (14, 13, and 13 respectively) and 2014 had the most approvals of the time period studied with a total of 19 NMEs and biologics (Fig. 1b). Pfizer, AstraZeneca, and Abbott/AbbVie had 100% of their launches as NMEs while Amgen, Eli Lilly and Roche all had about two-thirds of their approvals as biologics (Fig. 1a). Industry-wide, the percentage of approvals for biologics increased during the period of 2011-2016 (Fig. 1b). When the approvals are categorized by therapeutic area (Fig. 1c), oncology had the largest number (43) while ophthalmology and rare disease had the fewest (2). When observed by therapeutic area, immunology had the largest percentage of biologic launches approved (60%) and neuroscience the least (0%). The latter is not surprising given the challenge in large molecules accessing targets in the central nervous system.

### 3.2 Drug pre-approval publication trends

The number of research publications on drugs (not drug targets as in [21]) prior to their approval was used as a measure of research activity contributing to the drug launch. In total, there are 6781 research papers identified for the 113 unique approved drugs. Specifically, there are 1249 papers for 31 biologics and 5532 papers for 82 NMEs. The top 5 drugs as judged by pre-approval publication density are fingolimod hydrochloride, everolimus, tocilizumab, ivabradine, and sugammadex. Each of these 5 drugs had over 300 research papers published in advance of their approval date. There are also 4 drugs that did not have any prior published research including dronedarone hydrochloride, eltrombopag olamine, indacaterol maleate, and viekira pak; viekira pak is a fixed drug combination and our analysis does not include any publications on the individual drug components. On average, there are approximately 60 papers for each drug; 40 papers for each biologic and 67 papers for each NME.

Figure 2a) shows the publication activity preceding an approval, where the Y-axis is the publication count and the X-axis is the number of years preceding approval. Most publications occurred within 10 years of approval. For each publication on a drug, we calculated the years elapsed between the publication date and the approval date of the drug. The average year elapsed for both NME's and NBE's is around 2 years. For each drug in our dataset, we also calculated the time elapsed between the first research publication specifying the drug name and the approval date. The time from first publication to approval generally hovered around 6.3 years and again there was no appreciable difference between large and small molecules. In a few instances the research support dated back as early as 20 years prior. For example, belatacept (cytotoxic T-lymphocyte antigen 4 immunoglobulin) is a biologic that was approved in 2011 for preventing transplant kidney rejection. It has multiple papers on early stage research in cellular and animal models back to the mid-90s and early 2000 [22,23,24,25]. This is consistent with a recent finding that today's important medicines are often associated with fundamental discoveries that occurred decades before the approval [26]. That conclusion is slightly different than our overall analysis in which we required the publication to study the drug, and not just the drug target.

If we organize the papers by the company that launched the drugs, it highlights the overall research volume behind each company's approvals. We correlated the number of approvals with the total number of publications (i.e., equivalent to research volume) using Pearson coefficient (Figure 3). Across all companies, there is a strong correlation between the number of biologic approvals and the number of papers (correlation: 0.88) while there is moderate correlation between a company's total number of NME approvals and the number of papers (correlation: 0.71). For biologic approvals, Roche has the largest number of papers (616), followed by BMS (169) and Sanofi (151). For NME approvals, Novartis has by far the largest total number of papers (1607), followed by Merck (877) and Pfizer (571). While not shown, we also calculated the number of papers/drug approval for each of the 13 assessed companies. For biologics, Roche averages the most papers/drug (88) and Merck the fewest (4). For NME's, Amgen averages the highest number of papers/ drug (184) with GSK the fewest (27). In aggregate, the analysis demonstrates that there is a good correlation between the number of papers published in advance of drug approvals and the number of drug approvals across these 13 companies. But, it also shows that there are a few entities; *e.g.* Pfizer and GSK, with limited numbers of publications associated with their drug approvals as compared to their peers.

We next stratified the papers by therapeutic area. Figure 4 shows the total number of papers prior to approval compared with the number of approvals in each therapeutic area. For biologic approvals, immunology has the largest number of papers (553), followed by oncology (482) and endocrine (75). That immunology is the leader for biologics is consistent with many of these drugs being tested in multiple autoimmune disorders such as rheumatoid arthritis, psoriasis, and ulcerative colitis. The correlation between the number of biologic approvals and the number of papers is 0.81. For NME approvals, oncology has the largest number of papers (2309), followed by cardiovascular disease (828) and immunology (771). The correlation between the number of NME approvals and the number of papers is 0.93. While oncology had the total highest number of NMEs approved, it was still disproportionately high in the number of publications per approval relative to the other therapeutic areas. Given the feverish pace of oncology R&D across the industry with many immuno-oncology drugs in dozens or hundreds of combination trials, this result is not surprising.

### 3.3 Primary institutes/Contributors

In addition to total publication volume, we also analyzed the primary authoring institutes of the papers to measure contributions as a function of type of research organization. We classified the primary institutes into several types, including academics (e.g., universities, hospitals, medical institutes), top pharma, other pharma and biologic company, government agencies (e.g. NIH), and others (e.g., private practices). On average, academics contributed 79% of the pre-approval publications on biologic drugs and 76% on NMEs while top pharma publications account for only 10% for biologics and 13% for NMEs. Other pharma/ bio companies and government agencies contributed on average <5% of all publications.

Figure 5a details on a company- by-company basis the research contributors and the percentage of their contributions. For biologic approvals, academics contributed significantly (>50%) to the publication total of launched drugs for all companies. Top pharma's research contribution is less than 45% of the total contributions for each company. Not surprisingly, other pharma and biotech companies only made small research contributions - less than 7% in all cases. For NME approvals, academics again contributed significantly to the publication total of launched drugs with over 50% contributions for all companies. Top pharma's own research contribution is less than 29% of the total contributions across all companies.

Other pharma and biotech companies made more measurable contributions to approvals for NME's than biologics, especially in Takeda's case with 22% of contributions. This difference can be attributed to the limited number of companies with capabilities in developing biologic drugs during the 2006-2016 era.

For the biologic approvals, as delineated by therapeutic area (Fig. 5b), academics similarly contributed over 70% of the research volume in all therapeutic areas, (excluding infectious diseases where there are only 4 papers total). For NME approvals, academics again contributed significantly (over 50%) to the publication total of launched drugs in all therapeutic areas.

## 3.4 Research Focus

We next studied the research focus/expertise of the different contributors by analyzing the top 15 most frequent MeSH terms associated with their respective publications. Figure 6a shows the relative contributions of different organizations for each of the top 15 MeSH terms. For biologic approvals, academic institutes constitute the majority of publications for all MeSH topics except for 'pharmacokinetics'. They dominated the fields of physiology (92%), pathology (91%), adverse events (84%), therapeutic use (83%), and metabolism (81%). Top pharma tended to emphasize 'pharmacokinetics' in their publications more than other organizations with a 59% contribution, and they likewise have a high percentage of research in 'chemistry' (44%) and 'analogs and derivatives' (34%). Top pharma leading in pharmacokinetics and chemistry is consistent with the expertise of the industry. Other pharma and bio companies had relatively smaller percentages of research (less than 10%) for all topics. Other uncategorized contributors had a contribution range from 0% to 6%. Government agencies contributed a very small portion with a range from 0% to 5% as they support modest intramural research institutions with their primary goal to appropriately direct government research funding to the best academic groups who can use the funding for public good.

For NME approvals, as was the case with biologics, academic institutes have a higher percentage of research terms than other organization except for 'pharmacokinetics'. They dominated the fields of pathology (89%), adverse events (85%), immunology (83%), therapeutic use (82%), physiology (81%), and drug therapy (80%). Top pharma again dominated the field of 'pharmacokinetics' with a 53% contribution, and they have over 20% contribution in 'chemistry' (30%) and 'antagonists & inhibitors' (23%). Other pharma and bio companies had a contribution range from 2% (adverse events) to 8% (pharmacokinetics). Other uncategorized contributors had a contribution range from 2% to 7%. Government agencies contributed a small portion with a range from 1% to 3%.

## 3.5 Publications of the Sponsor

It was also of interest to evaluate whether companies varied in their approach to publishing on their own drugs, thus we also conducted an assessment of this subset of the publications. Novartis was the most prolific with 145 papers, followed by Merck and Pfizer with about 100 papers each. AbbVie published the least with only 6 papers. For each publication on a drug, we calculated the years elapsed between the publication date and the approval date of the drug. The average year elapsed for all drugs is around 0.72 years. This timeframe is shorter than that for all contributors averaged together (2 years).

Novartis, AstraZeneca and AbbVie tended to publish close to approval (0.4-0.6 years prior) while Pfizer was the earliest with 3.7 years from publications to approval.

We also analyzed the research focus of this subset of Sponsor publications by MeSH analysis. Their top 15 most frequent MeSH terms are pharmacology, drug therapy, antagonists and inhibitors, pharmacokinetics, drug effects, analogs and derivatives, metabolism, therapeutic use, administration and dosage, chemistry, genetics, blood, immunology, prevention and control, and physiology. These MeSH terms, such as pharmacokinetics, are highly related to the expertise of pharma and are critical elements of data packages that support NDA filings. In contrast, we have determined that academic contributions are more heavily weighted towards pharmacology/physiology which highlights that the compendium of data provided by pharma and academia provides a comprehensive public view of the data for approved drugs.

### 3.6 Analysis of failed drugs

As a comparator, we also examined pre-approval publication history for drugs that terminated development in Phase 3. In total, there are 524 research papers associated with the 39 failed drugs. The top 5 drugs as rated by number of papers are astuprotimut-r, almorexant, etec vaccine, acadesine, and amibegron with each of them having 45 or more papers published prior to their latest termination date. On average, there are approximately 13 papers per drug: 18 papers for each biologic and 12 papers for each NME. This contrasts greatly with approved drugs, which have approximately 60 papers per drug (40 papers for each biologic and 67 papers for each NME).

Figure 2b shows the publication activity preceding a termination, where the Y-axis is the publication count and the X-axis is the number of years preceding termination. For each publication on a drug, we calculated the year elapsed between the publication date and the latest termination date of the drug. The average year elapsed for all drugs is around 5.1 years. For each drug in our dataset, we also calculated the time elapsed between the first research publication specifying the drug name and the termination date. The time from first publication to termination generally hovered around 8.2 years. In both instances, this timeframe is longer than for approved drugs.

Using the same criteria as for approved drugs, we analyzed the primary authoring institutes of the papers to measure contributions by different organizations. On average, academics contributed 72% of the pre-termination publications on biologic drugs and 60% on NMEs. On the other hand, top pharma publications account for 7% for biologics and 19% for NMEs. As was true for approved drugs, the % contributed by other pharma/bio companies or government agencies constituted a small minority of the total publications.

We also studied the research expertise of the different contributors by analyzing the top 15 most frequent MeSH terms associated with their respective publications and compared it to the approved drugs as shown in Figure 6b. For both approved and failed drugs, academic institutes make the largest contribution to most of the research topics but their contribution dropped from 76% to 65% when averaged across NMEs and biologics. The 13 top pharma have similar publication percentages (13% for approved and 14% for failed drugs). Conversely, smaller pharma and biotechs contribute to a greater % of the publications on failed drugs (11%) than they do approved drugs (5%).

# 4. Discussion

Our work was undertaken with the goal to better understand the potential relationship between publication history (e.g. affiliation and focus of the science) prior to FDA approval of large and small molecule drugs. We conducted this analysis for the drugs from the top 13 pharma that received FDA approval between 2006-2016 which accounts for approximately one-third of NME and half of the NBE products taken to market in the US during that decade. While we focused on successful drug approvals because our goal was to study factors that are associated with success as defined by FDA approval, we also sought to look for differences (if any) for those drugs that were terminated by these companies during that same timeframe.

On average, we determined that there are about 60 original research papers supporting each approval; 40 papers supporting each biologic and 67 papers supporting each NME approval. For both, there was a reasonable correlation between total number of prior approval publications and total number of drug approvals. This was also true when broken out by therapeutic area although oncology drugs had a disproportionately high number of publications relative to approvals. A few companies differentiated in having many fewer publications/drug approved. Determining why this is so (*e.g.* company publication policy, something unique to the targets that limited publishing by non-company entities, *etc.*) is beyond the scope of our analyses. We did, however, determine that the timeframe between first publication and drug approval did not differ for these companies so it is not due to a timing distinction. As a comparator, we also evaluated publication history for 39 candidate drugs that failed in Phase 3. There are approximately only 13 papers for each of these molecules; 18 papers for each biologic and 12 papers for each NME which is significantly less than for approved drugs. As the regulatory review time is generally only about 1 year for first time approvals, the significantly smaller volume of publications for terminated molecules is likely not attributable to simply a difference in timeframe of publication. Rather, it may suggest that approved drugs are often associated with a more robust data set provided by a large number of institutions.

To further interrogate the potential value of pre-approval publications to drug approvals we first analyzed the type of research contributors as judged by primary authoring institute. This is because 1) focusing on the primary institutes is a convention in the literature [12], and 2) our primary data source (PubMed) does not always list the affiliation of every author thus we lack the ability to track every non-primary institute on the papers. We grouped universities, university-affiliated hospitals, biomedical institutes (e.g., The Scripps Research Institute) into one category of 'academic' which contributed significantly to the publication total of launched drugs constituting 79% of the contributions for biologics and 76% of the contributions for NMEs. This major contribution by 'academics' held true for all companies and across all therapeutic areas. Conversely, top pharma published only 10% of the papers for biologic approvals and 13% for NME approvals while all other institute types contributed 5% or less of the publications for biologics and NMEs.

For failed drugs, academics contributed 72% of the pre-termination publications on biologic drugs and 60% on NMEs. The slightly lower % as compared to approved NME's was largely made up by top pharma and other pharma/bio company contributions with 19% and 16% for NMEs, respectively.

The research focus of the published literature showed breadth and diversity as determined using MeSH headings analysis of the entire publication dataset. Academics had a high percentage of research in 'physiology', 'pathology' and 'adverse events' for biologics; and a high percentage in 'pathology', 'adverse events' and 'immunology' for NMEs. Top pharma had a relative high percentage of research in 'pharmacokinetics', 'chemistry' and 'analogs and derivatives' for biologics; and a high percentage in 'pharmacokinetics', 'chemistry' and 'antagonists & inhibitors' for NMEs.  This is somewhat in contrast to reference [5], which concluded that academia mainly has strength (as judged by publication types), in basic research, target identification and target validation. Our analysis, directed to drugs rather than drug targets, suggests an even broader role for these organizations in providing key data on candidate drugs. One difference between our analysis and that of Reference [5] is that the previous work focused on 'subjective' descriptions of external innovation models while we have used text-mining tools that provided us detailed data and analysis support.

The MeSH analysis for failed drugs was consistent with approved drugs in that academic institutes make the largest contribution across most research topics, albeit their percent contribution decreased from 76% to 65%. On the other hand, small pharma and biotechnology companies contributed more to the overall research topics of failed drugs than in the case of approved drugs. While the reason(s) for this distinction cannot be ascertained by our analysis, it may suggest that these failed drugs, and their associated targets, were challenging but innovative targets that attracted the attention of smaller R&D organizations.

We conclude that our findings highlight the mutually beneficial strengths of the research organizations that publish their research findings on molecules that go on to become approved drugs.  It is consistent with the notion that full characterization of a drug candidate (target biology, efficacy, safety, pharmacokinetics, *etc*.) benefits from a combination of effort from academia and industry, and that academia is the source of the majority of these publications.  Our analysis supports the hypothesis that increased investment in the external innovation environment, where the expertise and capabilities of top pharma, academics, and other pharma/bio companies complement each other, is likely to manifest in a more effective discovery process and approval of innovative medications.

# Acknowledgements

We thank Dr. Kurt Rasmussen for reviewing the manuscript and making constructive comments.

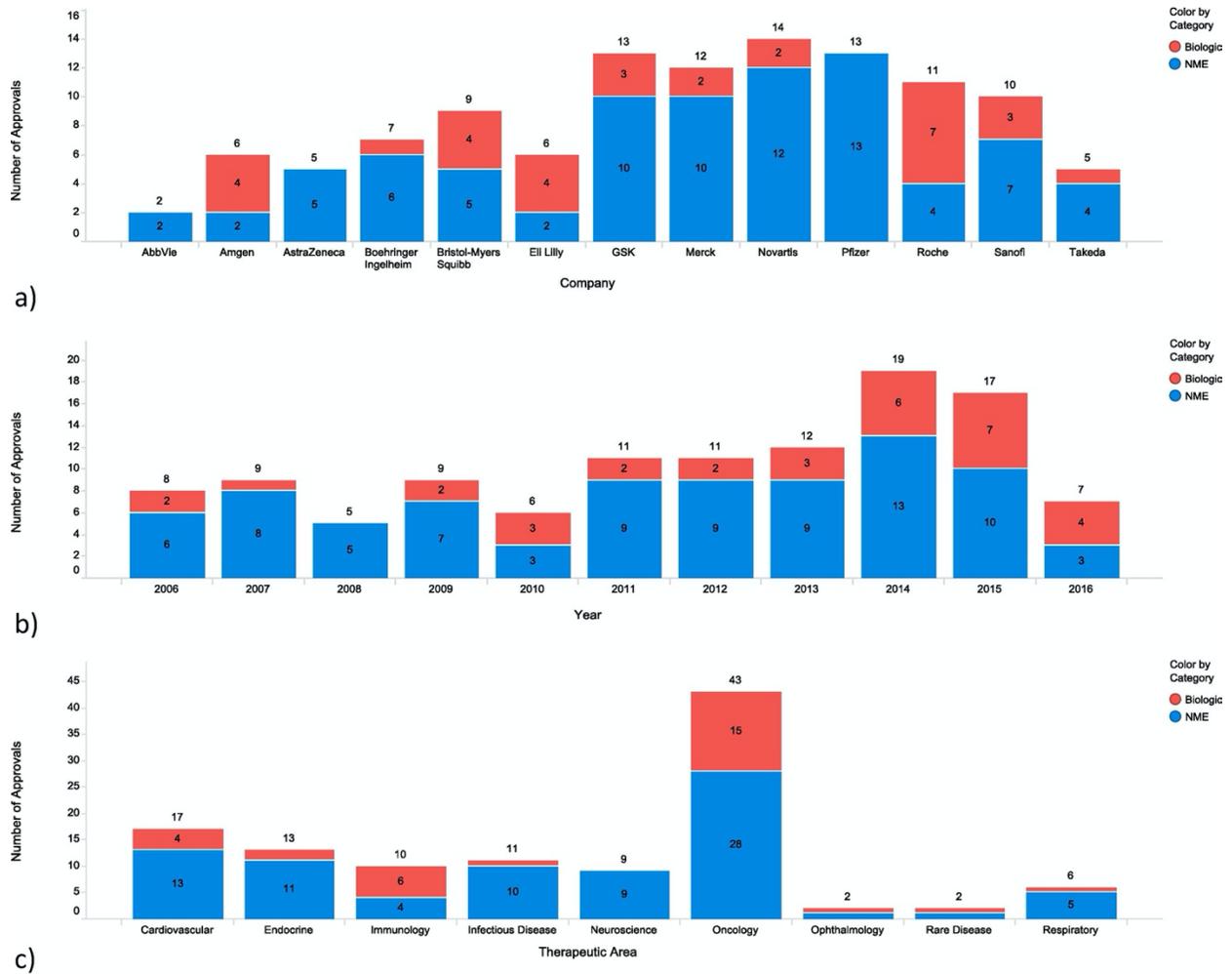

Figure 1. FDA approvals of 13 top pharma during 2006-2016. (a) Approval by company. (b) Approval by year. (c) Approval by therapeutic area.

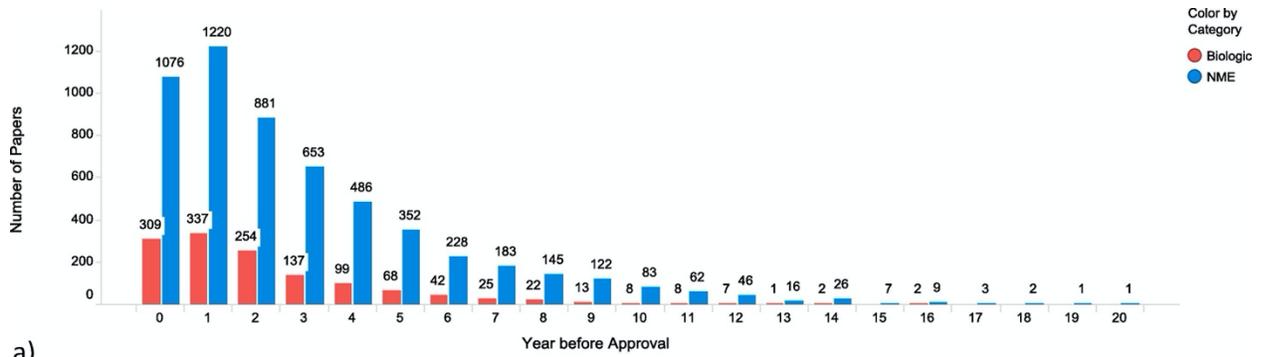

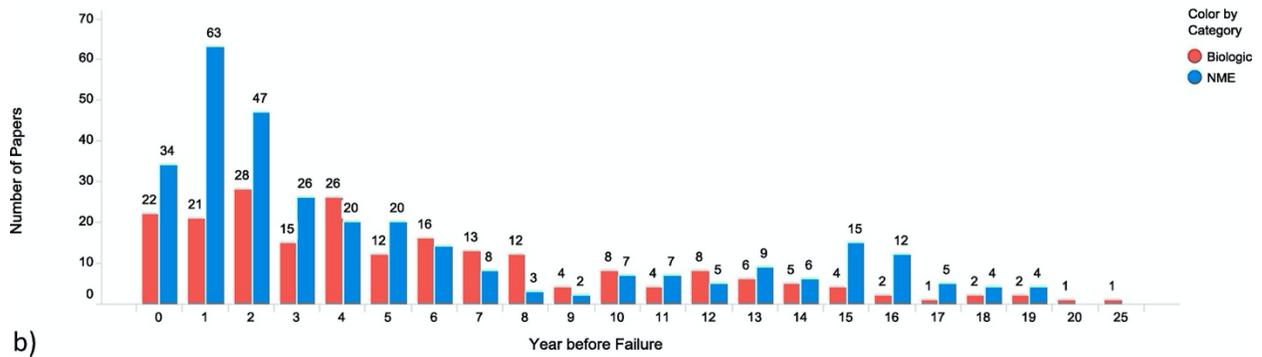

Figure 2. Publication history prior to a) Approval or b) Phase 3 termination for 13 top pharma in 2006-2016.

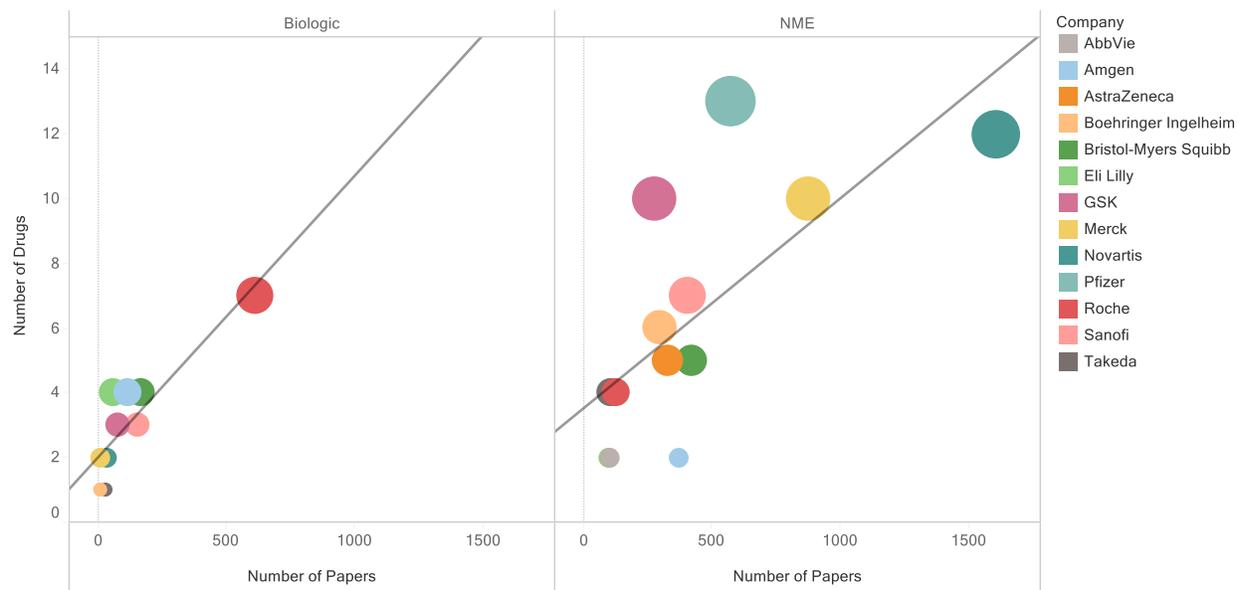

Figure 3. The number of papers (research volume) supporting each company's drug approvals. Bubble sized by the number of approvals. (a) The number of papers by biologic approvals. (b) The number of papers by NME approvals.

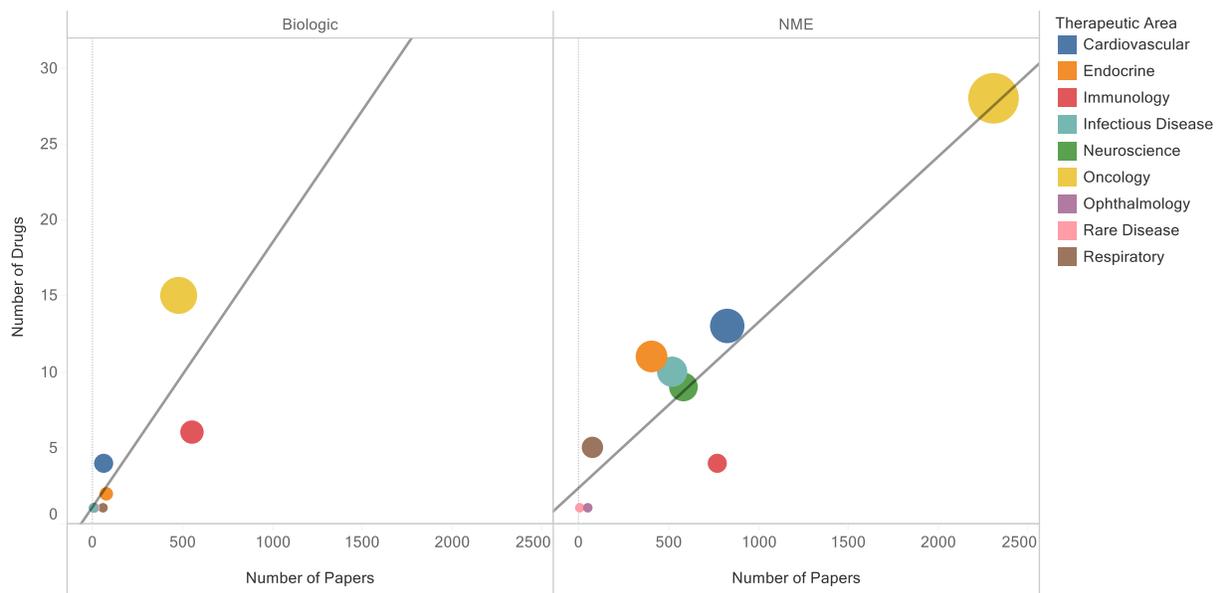

Figure 4. The number of papers supporting approvals across the therapeutic areas. Bubble sized by the number of approvals. (a) The number of papers by biological approvals. (b) The number of papers by NME approvals.

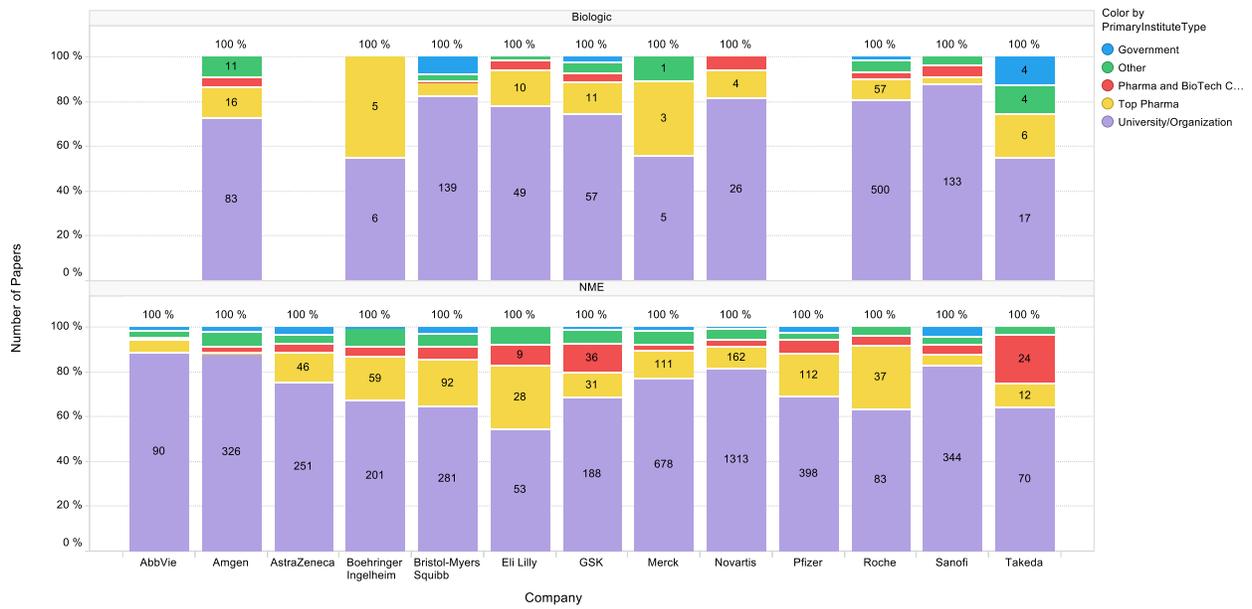
a)

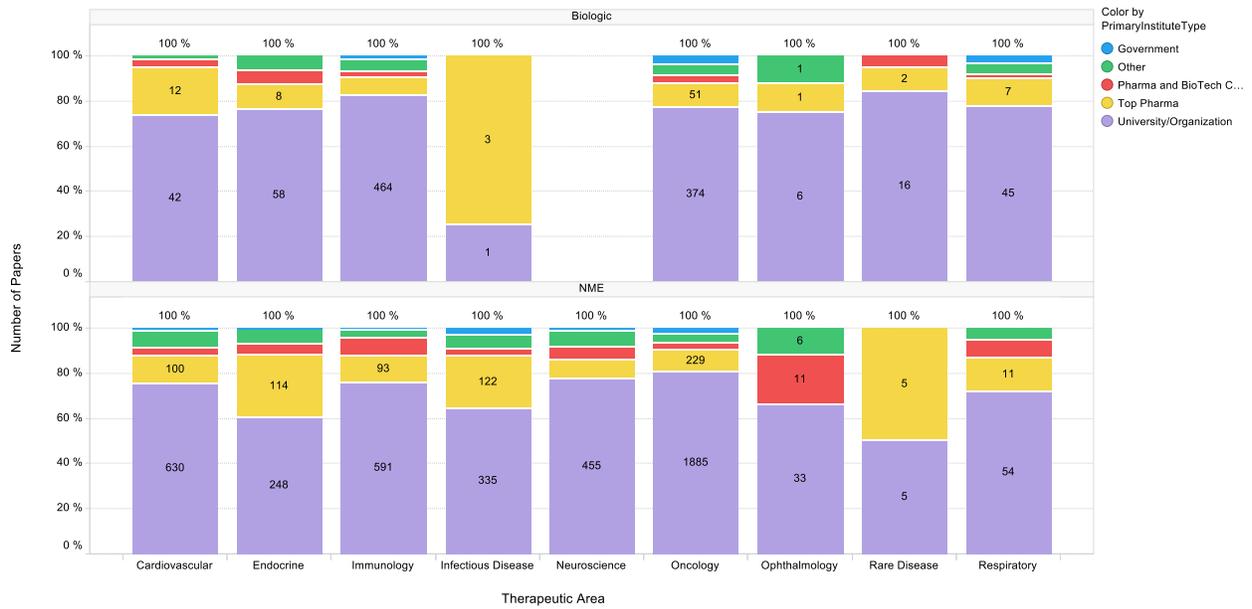
b)

Figure 5. Groupings of author affiliations for papers published on approved drugs. (a) All papers separated into individual companies. (b) All papers separated by therapeutic area.

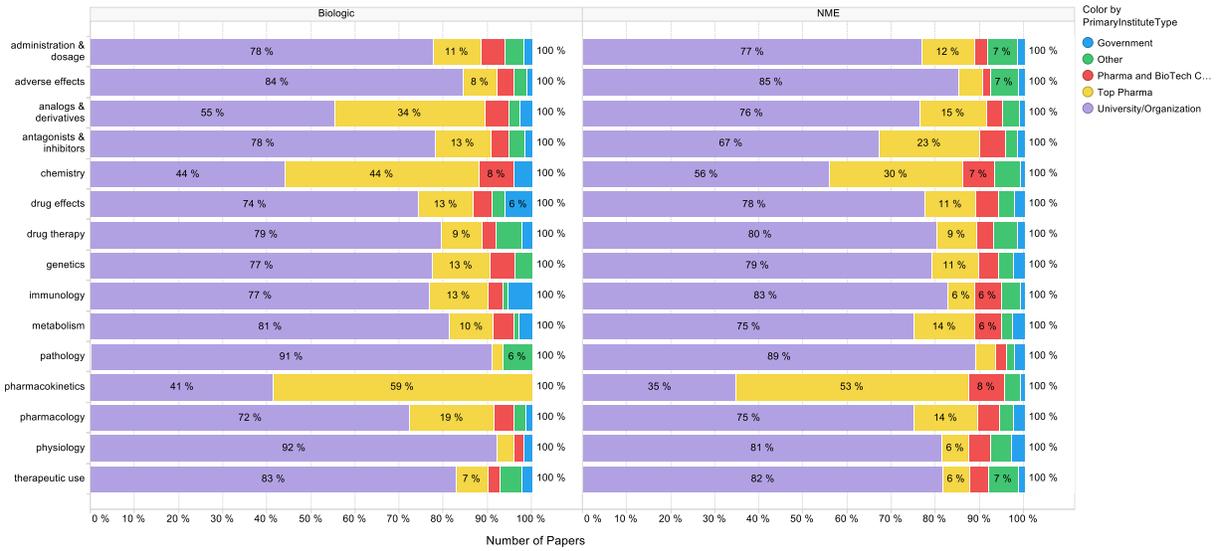

a)

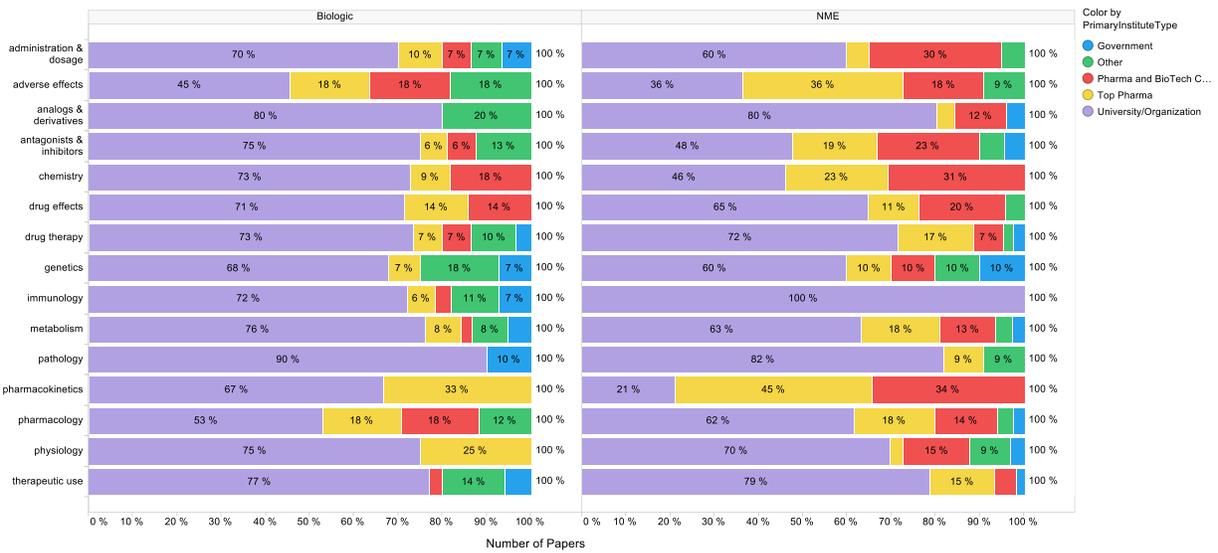

b)

Figure 6. The distribution of research topics represented by top 15 MeSH terms. The left panel shows the distribution for biologic launches while on the right is shown the distribution for small molecule launches. (a) MeSH analysis of the publications on approved drugs during 2006-2016. (b) MeSH analysis of the publications on failed drugs during 2006-2016.

18